# Brain tumour segmentation with incomplete imaging data

James K. Ruffle FRCR MSc[1], Samia Mohinta MSc[1], Robert Gray PhD[1], Harpreet Hyare FRCR PhD[1], and Parashkev Nachev FRCP PhD[1]

[1]Queen Square Institute of Neurology, University College London, London WC1N 3BG, UK

Correspondence to:

Dr James K Ruffle

Email: j.ruffle@ucl.ac.uk

Address: Institute of Neurology, UCL, London WC1N 3BG, UK

Correspondence may also be addressed to:

Professor Parashkev Nachev

Email: p.nachev@ucl.ac.uk

Address: Institute of Neurology, UCL, London WC1N 3BG, UK

**Key words:** Tumour segmentation; Neuroradiology; Magnetic Resonance Imaging; Deep Learning; Artificial Intelligence.

# Abstract

Progress in neuro-oncology is increasingly recognized to be obstructed by the marked heterogeneity—genetic, pathological, and clinical—of brain tumours. If the treatment susceptibilities and outcomes of individual patients differ widely, determined by the interactions of many multimodal characteristics, then large-scale, fully-inclusive, richly phenotyped data—including imaging—will be needed to predict them at the individual level. Such data can realistically be acquired only in the routine clinical stream, where its quality is inevitably degraded by the constraints of real-world clinical care. Although contemporary machine learning could theoretically provide a solution to this task, especially in the domain of imaging, its ability to cope with realistic, incomplete, low-quality data is yet to be determined. In the largest and most comprehensive study of its kind, applying state-of-the-art brain tumour segmentation models to large scale, multi-site MR imaging data of 1251 individuals, here we quantify the comparative fidelity of automated segmentation models drawn from MR data replicating the various levels of completeness observed in real life. We demonstrate that models trained on incomplete data can segment lesions very well, often equivalently to those trained on the full completement of images, exhibiting Dice coefficients of 0.907 (single sequence) to 0.945 (complete set) for whole tumours, and 0.701 (single sequence) to 0.891 (complete set) for component tissue types. This finding opens the door both to the application of segmentation models to large-scale historical data, for the purpose of building treatment and outcome predictive models, and their application to real-world clinical care. We further ascertain that segmentation models can accurately detect enhancing tumour in the absence of contrast-enhancing imaging, quantifying the burden of enhancing tumour with an $R^2 > 0.97$, varying negligibly with lesion morphology. Such models can quantify enhancing tumour without the administration of intravenous contrast, inviting a revision of the notion of tumour enhancement if the same information can be extracted without contrast-enhanced imaging. Our analysis includes validation on a heterogeneous, real-world 50 patient sample of brain tumour imaging acquired over the last 15 years at our tertiary centre, demonstrating maintained accuracy even on non-isotropic MRI acquisitions, or even on complex post-operative imaging with tumour recurrence. This work substantially extends the translational opportunity for quantitative analysis to clinical situations where the full complement of sequences is not available, and potentially enables the characterisation of contrast-enhanced regions where contrast administration is infeasible or undesirable.

## Abbreviated summary

Brain tumour segmentation models with incomplete sets of MRI sequences – common in clinical practice – still delineate lesions well and even identify enhancing tumour without post-contrast imaging. The observed small marginal benefits of additional MR sequences, especially contrast-enhanced, suggest the cost and risk to the patient of imaging sufficient to permit high-definition quantitative analysis may be reducible.

# Introduction

Progress in neuro-oncology is increasingly recognized to be obstructed by the marked *heterogeneity*—genetic, pathological, and clinical—of brain tumours. If the treatment susceptibilities and outcomes of individual patients differ widely, determined by the interactions of many multimodal characteristics[1], then large-scale, fully-inclusive, richly phenotyped data—including imaging—will be needed to predict them at the individual level. Such data can realistically be acquired only in the routine clinical stream, where its quality is inevitably degraded by the constraints of real-world clinical care. Although contemporary machine learning could theoretically provide a solution to this task, especially in the domain of imaging, its ability to cope with realistic, incomplete, low-quality data is yet to be determined.

Over the last few decades, lesion segmentation has formed a cornerstone of innovation across the domains of neuro-oncology[2-4], medical imaging[5,6], biomedical engineering[7], machine and deep learning[8]. The ability to segment an anatomical or pathological lesion in three-dimensions confers the ability to evaluate it quantitatively – moving beyond visual qualitative assessment – with greater richness and fidelity than conventional two-dimensional measurements repeatedly shown to be often spurious and inconsistent between radiologists[9-11], and with greater sensitivity to the heterogeneity of the underlying pathological patterns[12]. Enabling radiological image segmentation opens a wide array of possibilities for downstream innovation in neuro-oncological healthcare and research, ranging from clinical stratification, outcome prediction, response assessment, treatment allocation and risk quantification, many of which have already shown great promise. The underlying goal is to enhance the individual fidelity of data-driven decision making, facilitating better patient-centred care[13-15], a remit especially warranted in neuro-oncology.

The segmentation of brain tumours remains a particularly challenging task owing to the marked heterogeneity of their imaging appearances: spatial distribution, morphology, signal characteristics, and impact on adjacent healthy anatomical structures[16-18]. Its difficulty has even inspired an international competition for cutting-edge deep learning groups to create the best segmentation model. Known as the Brain Tumour Segmentation challenge (BraTS), it is attracting increasing attention as well as support from both the Radiological Society of North America and the American Society of Neuroradiology, providing large scale data with multimodal MRI – FLAIR, T1, T2 and contrast-enhanced T1 (T1CE) sequences – as well as the labelled ground-truths of oedema, non-enhancing and enhancing tumour[8,19,20].

But while benchmark tasks have unquestionably aided the advancement of lesion segmentation – indeed of computer vision generally – they have compelled a research focus on developing uniformly multimodal models trained on sequence-complete acquisition sets, often rare in real-world clinical practice. The causes of incomplete data are legion, but common examples include patient contraindications to contrast, corruption by image artefacts, and image acquisition constraints such as those imposed in pre-operative stealth studies. Taking just one of many possible causes for image degradation, the prevalence of motion artefact has been reported as 7.5% of outpatient and 29.4% of inpatient MRI studies, with an estimated economic impact of $115,000 per scanner, per year[21].

The real-world utility of tumour segmentation must lie within the clinical domain, such as for treatment planning and monitoring across neuro-oncology. Yet, the ability to undertake segmentation in these real-world clinical situations, where complete - *'perfect'* - data is scarce, remains completely unknown. How well do contemporary segmentation modelling architectures perform when trained on sequence-incomplete data, and what features of the lesion are correctly identifiable under such circumstances?

Here, we aimed to systematically quantify and answer these questions with the largest and most comprehensive study of its kind based on the application of state-of-the-art deep learning tumour segmentation models to large-scale MR imaging of brain tumours. We hypothesized that the decrement in segmentation performance with the loss of sequences would be modest, rendering good quality segmentations feasible with incomplete data.

## Materials and methods

### Data

The study was approved by the local ethics committee. We received ethical permission for the consentless analysis of irrevocably anonymized data collected during routine clinical care.

We used the BraTS 2021 challenge data for all model training. This dataset is described in detail by its curators elsewhere[20,22,23]. In brief, it includes a large retrospective sample of multi-institutional brain tumour MRI scans, with heterogeneous equipment, protocols, and image quality. The following sequences are included: T1-weighted, T2-weighted, FLAIR and post-contrast T1 (T1CE), with a pre-processing pipeline consisting of image co-registration, sampling to a 1mm$^3$ isotropic space and skull-stripping. Lesions were segmented with an ensemble of previous top-ranking BraTS algorithms with subsequent manual refinement and checking by a panel of board-certified attending neuroradiologists with more than 15 years of clinical experience in neuro-oncology[24]. We used the training set of 1251 individuals of the BraTS 2021 challenge data—comprising 5004 separate images—as this group included all ground-truth labels for model cross-validation.

Having trained and evaluated a set of models on the BraTS 2021 challenge data, we sought to separately evaluate their performance on an additional held-out population from our own centre. The aim of this was to provide an additional robust safeguard of model performance with international and external validation. Specifically, we acquired retrospective imaging for a random sample of 50 individuals who underwent gadolinium-enhanced MRI head studies between 2006 and 2021 for a known glioblastoma as part of their routine clinical care at our centre. The random allocation of year selected was to further instil heterogeneity to our sample, as data would be acquired over one of 11 possible MRI scanners of both 1.5 (n=5) and 3T (n=6) field strengths, from multiple different manufacturers, and over a 15-year period.

Moreover, of our 50 participants, we also chose to include 10 of those with post-operative imaging and evidential tumour recurrence. This choice increased the difficulty of the task, for a model would need to recognize post-operative resection/surgical bed as separate from the subsequent disease recurrence, as well as capturing the instrumental heterogeneity of different MRI machines distributed in time and place.

Most of our sample did not include volumetric imaging, a reflection of local clinical practice at the time of acquisition. To improve harmonization, we therefore employed super-resolution in the processing pipeline[25,26]. The pipeline yielded data in a similar format to the BraTS challenge data[20] with 1mm$^3$ isotropic and skull-stripped multi-sequence data. Lesions were hand-labelled with ITK-SNAP by a neuroradiology fellow with 3 years of experience working with brain tumour imaging, with additional aids of the ITK-SNAP semi-automated segmentation tools, namely Random Forest based classifiers with subsequent manual refinement[27].

Tumour annotations conform to established tissue class labels comprising gadolinium-enhancing tumour, peritumoural oedema/invaded tissue and non-enhancing tumour/necrotic tumour core[19]. The detailed description of these components is beyond the scope of this article and is discussed elsewhere[19,20]. In brief, enhancing tumour refers to regions with visible enhancement on a T1CE sequence after gadolinium administration. Non-enhancing tumour/necrotic tumour core refers to the part of the tumour that does not enhance after gadolinium, typically deep to the enhancement, while oedema/invaded tissue refers to the peritumoural oedematous and/or infiltrated brain parenchyma, typified by hyperintensity on T2 and FLAIR sequences. Examples of these lesion tissue compartments as shown throughout Figures 2-5, and Figure 8.

## Algorithm

Our task was not to propose a new architecture superior to those already evidenced by the BraTS 2021 challenge. Rather, we sought to characterize, evaluate, and quantify the variation in model performance with increasingly incomplete data, as a proxy index of translational potential across the variety of clinical situations where full complete datasets rarely occur. We chose the nnU-Net self-configuring deep learning biomedical image segmentation modelling architecture[28], which notably won both the medical segmentation decathlon and the 2020 BraTS challenge[29,30]. In brief, this segmentation method is able to automatically configure itself, including in pre-processing, architecture, training and post-processing across any task, and has been shown to be a superior methodology across a range of public datasets and tasks, including brain tumour segmentation[28]. Our choice was guided by its excellent performance, and the simple, largely automated processing and training cycle, which made development across many models at scale feasible.

Each nnU-Net[28] is in particular a self-configuring U-Net[31], incorporating the standard encoder-decoder architecture and skip connections, instance normalization and leaky rectified linear units. We used the 3D architectural formulation in all experiments. The nnU-Net approach employs a polynomially decaying learning rate, initially set to 0.01, with stochastic gradient descent optimization. The loss function is a weighted sum of the Sørenson-Dice coefficient and cross-entropy. Training data is augmented on the fly, including with rotations, scaling, Gaussian noise and blur, brightness and contrast shifting and gamma correction. Patch and batch size are also self-configured. Model training utilizes 1000 epochs, with foreground oversampling to mitigate the impact of class imbalances. We used 5-fold cross-validation for each experiment and its evaluation with the BraTS 2021 challenge data, as well as additional external/international out-of-sample evaluation of models with the additional data from our own centre as detailed above. A schematic of the model architecture is shown in Supplementary Figure 1.

### Statistical analysis and performance evaluation

We trained all possible combinations of the MRI sequences T1, T2, FLAIR and T1CE as separate models. This included all models using only a single sequence, two sequences, three sequences and finally a complete four-sequence model. We also trained separate models for abnormality detection (i.e., a binary lesion mask to detect and segment the whole tumour) as well as tumour segmentation with the tissue classes of oedema, enhancing and non-enhancing tumour. This approach comprised 30 different models in total.

Performance was principally quantified by the out-of-sample Sørenson-Dice coefficient between ground truth and inferred labels[32,33], in accordance with typical research practices[3,8]. This metric derives the area of overlap between the model prediction and the labelled ground-truth. The Sørenson-Dice coefficient, or Dice coefficient, is given as:

$$Dice = \frac{2(TP)}{(TP + FP) + (TP + FN)}.$$

We also quantified overall model accuracy, false discovery rate, false negative rate, false omission rate, false positive rate, negative predictive value, precision, and recall, ensuring a broad range of possible performance metrics[34]. All listed metrics were derived for whole tumour and the separate tissue constituents of oedema, enhancing and non-enhancing tumour, including with 95% confidence intervals (CIs), which are provided in detail throughout the supplementary material.

We constructed regression models between ground truth tumour volumes and model predictions, reporting the $R^2$. We acquired the acquisition times of contemporaneous imaging protocols at our centre for a given imaging sequence, to allow comparison between a gain in model performance aligned to the time it would take to be acquired. Lastly, we applied t-distributed stochastic neighbour embedding (tSNE)[35] - a nonlinear dimensionality reduction technique – to the contrast-enhancing components of all lesions in the BraTS dataset to create a two-dimensional representation of the lesions, projecting their high-dimensional similarities and differences into a readily surveyable space. We overlaid lesion volume and the Sørenson-Dice coefficient of lesion segmentations to display any variation in these indices with the morphology of the lesion.

### Data and code availability

All BraTS 2021 challenge data is readily available from the challenge website here: http://braintumorsegmentation.org[8,20]. Modelling code is readily available from the nnU-Net authors here: https://github.com/MIC-DKFZ/nnUNet[28]. All trained model weights are available upon request. Patient imaging data from our external validation site is not available for dissemination under the ethical framework that governs its use.

### Compute

All models were trained on an NVIDIA DGX-1 with 8 16GB Tesla P100 GPUs. With approximately 3.5 days to train a single model, the task required just over 13 days utilization of all cards.

## Results

### Incremental performance with sequence addition

All models performed well on whole tumour segmentation qualitatively, despite varying degrees sequence-completeness, with quantitative performance ranging from a Dice coefficient of 0.907 (95% CI 0.904-0.910) (single sequence) to 0.945 (95% CI 0.943-0.947) (complete sequence set) (Figure 1). Results for segmentation of the oedema, enhancing, and non-enhancing components were more variable, with Dice coefficients ranging from 0.701 (95% CI 0.689-0.713) (single sequence [FLAIR] segmenting non-enhancing tumour) to 0.891 (95% CI 0.886-0.896) (complete sequence set [T1+T2+FLAIR+T1CE segmenting oedema]). Of note, the models that performed the poorest typically struggled in the segmentation of the non-enhancing tumour component, particularly affecting single sequence models of T1, T2 and FLAIR, two and three-sequence models employing combinations of the former (i.e., with the omission of contrast). There was no evidence of model over-fitting when reviewing the training/validation curves. We provide the full breakdown of Dice coefficients for all models in Figure 1. Example image segmentations across the range of all models are provided in Figure 2 and 3, which visually illustrate excellent coverage of the lesion by the models, with relatively little error. We additionally detail model accuracy, false discovery rate, false negative rate, false omission rate, false positive rate, negative predictive value, precision, and recall (all with 95% CIs), for whole tumour and the separate tissue constituents of oedema, enhancing and non-enhancing tumour, all of which is provided within the supplementary material.

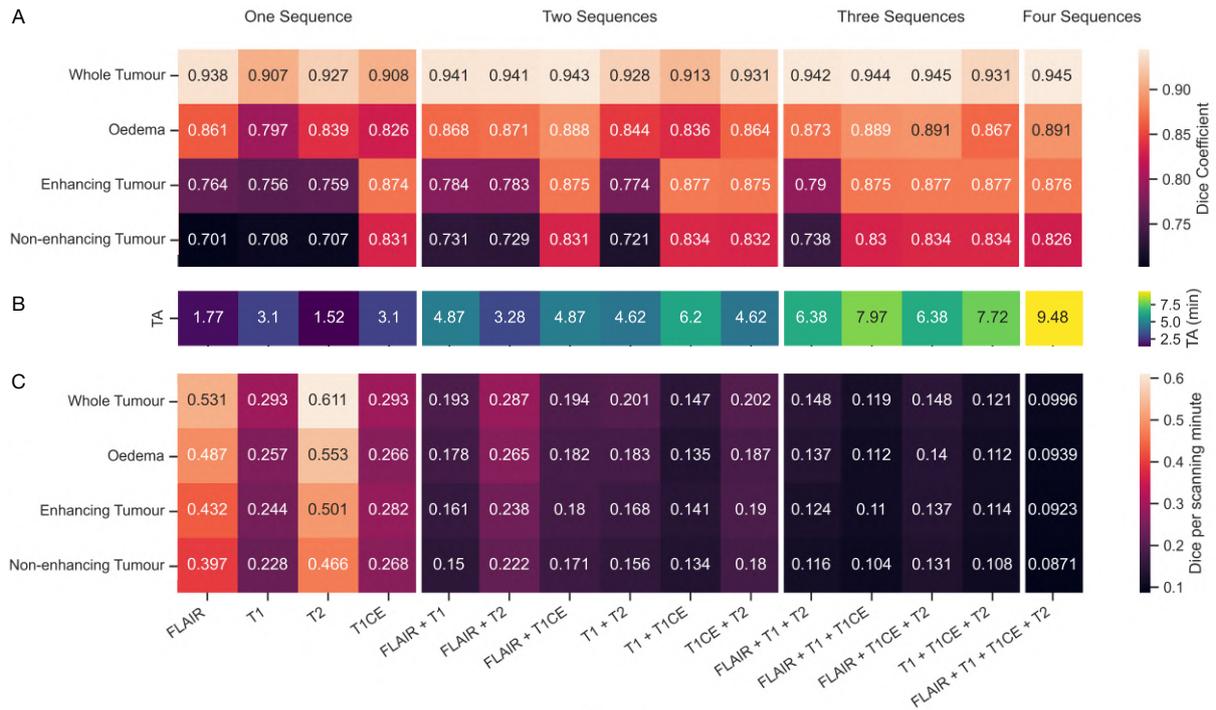

Figure 1: **Performance of all model combinations**. A) Heatmap illustrates the validation Dice coefficient across all models, for both whole tumour and the individual components. Models are partitioned into those which utilized just one sequence, two, three and finally the complete four-sequence model. A brighter orange/white box depicts a better performing model as per the Dice coefficient. B) Second heatmap depicts the relative acquisition time (TA) (in minutes) for the sequences used for a given model, with a more green/yellow box illustrating a longer acquisition time. C) Third heatmap illustrates the performance gain in Dice coefficient per minute of acquisition time. The mathematical derivation of the Dice coefficient is given in the methods. Colour keys are given at the right of the plot.

### Trade-off between acquisition time and segmentation fidelity

We aligned the acquisition times of all possible combinations of sequences using contemporaneous scanner protocol data at our centre, and from which determined the gain in model fidelity in Dice per scanning minute (Figure 1). This demonstrated that certain combinations of sequences appeared to offer greater gains in segmentation performance when compared with others, offering an insight into the efficiency of data acquisition in this clinical context. For instance, it was noted that whilst a single volumetric T1CE acquisition (proxy for a contrast-enhanced MRI stealth study for neurosurgical planning) took 3.1 minutes, achieving a whole tumour Dice coefficient of 0.908 and reasonable performance on individual components (see Figure 1), the addition of FLAIR raised total scanning time to only 4.9 minutes while improving whole tumour Dice to 0.943, just below the best performing model with all four sequences (Dice coefficient 0.945). Similarly, the three-sequence acquisition of FLAIR + T1CE + T2 (i.e., neglecting the pre-contrast T1) achieved Dice coefficients for whole tumour segmentation essentially equivalent to that of the complete four sequences, and reduced scanning time by 33%, from 9.48 to 6.38 minutes. We do, of course, note the omission of a pre-contrast T1 brings its own issues in delineating contrast from, for example, haemorrhage, but is nonetheless a striking illustration of how models with incomplete data still achieved comparable performance.

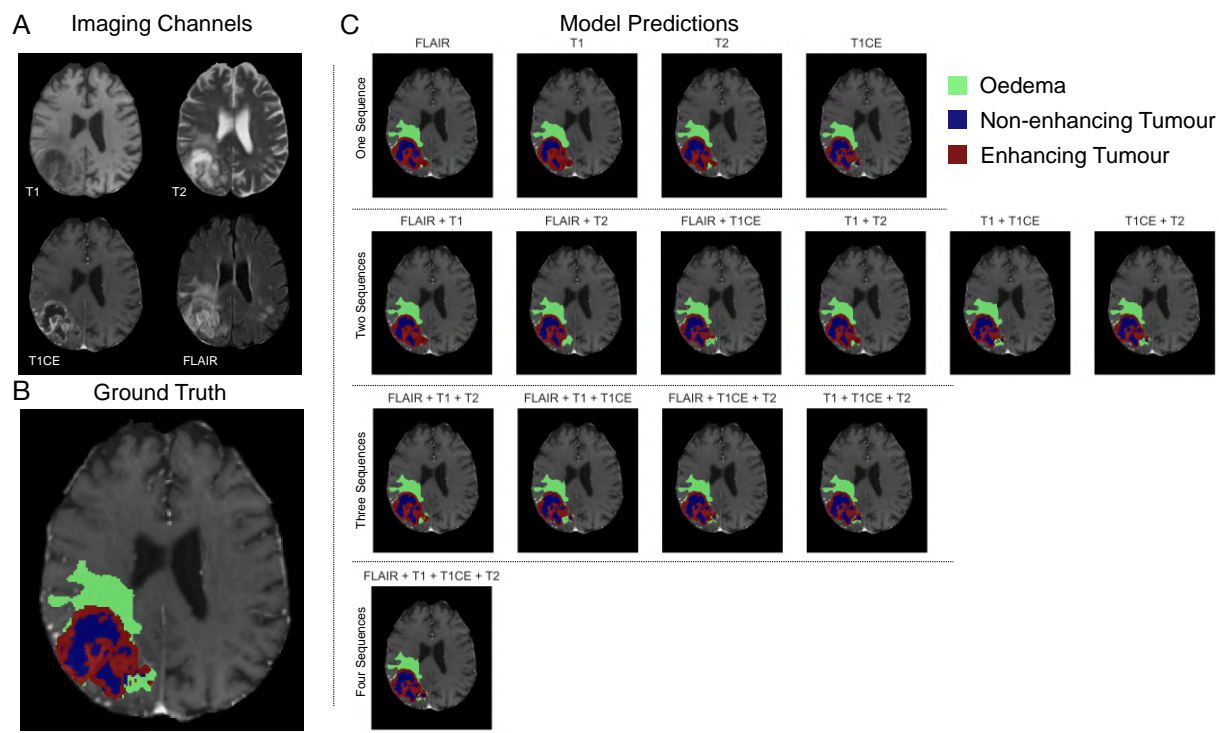

Figure 2: Example segmentation results. A) Left upper panel illustrates stacked axial slices of a given lesion for all imaging sequences, with B) corresponding radiologist-labelled ground-truth in the left lower panel. C) Right panel illustrates the tumour segmentation predictions across all model formulations, aligned to the number of sequences supplied.

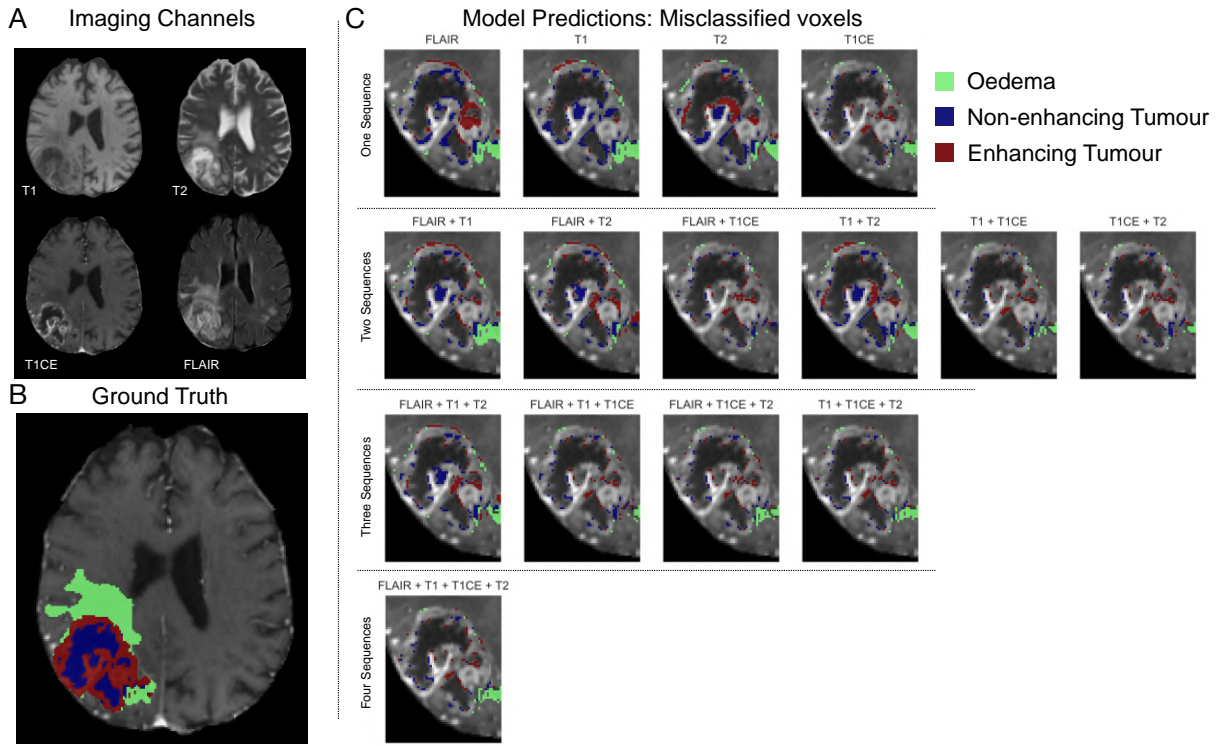

Figure 3: Minimal error in example segmentation results. A) Left upper panel illustrates stacked axial slices of a given lesion for all imaging sequences, with B) corresponding radiologist-labelled ground-truth in the left lower panel. C) Right panel illustrates the tissue-specific error in tumour segmentation predictions across all model formulations, aligned to the number of sequences supplied.

### Segmenting enhancing tumour without contrast-enhanced imaging

Interestingly, we discovered that models without contrast-enhanced imaging could still delineate tumours relatively well (Figure 4-5). Models without contrast imaging segmented whole tumour lesions with Dice coefficients ranging from 0.907 (95% CI 0.904-0.910) (single sequence – T1) to 0.942 (95% CI 0.940-0.945) (three sequences - FLAIR + T1 + T2). Of note, this latter performance was only just shy of the best performing full four sequence model with Dice of 0.945 (95% CI 0.943-0.947). Furthermore, models without the contrast-enhanced T1 sequence could still identify the enhancing tumour component well, with Dice coefficients ranging from 0.756 (95% CI 0.748-0.765) (single sequence – T1) to 0.790 (95% CI 0.782-0.798) (three sequences - FLAIR + T1 + T2) (Figure 4-5). This included the model's ability to identify and segment lesions where the focus of enhancing tumour was less than 7mm in diameter (Figure 5). The volume of enhancing tumour was highly significantly correlated to that of all model predictions, even despite contrast-enhanced imaging not being provided. The relationship between actual enhancing tumour volume to that of the model predictions with the following inputs were as follows: FLAIR alone ($R^2$ 0.964); T1 alone ($R^2$ 0.953); T2 alone ($R^2$ 0.966); FLAIR + T1 ($R^2$ 0.973); FLAIR + T2 ($R^2$ 0.976); T1 + T2 ($R^2$ 0.962); FLAIR + T1 + T2 ($R^2$ 0.972). Furthermore, inspection of the t-SNE-derived low dimensional representation of the lesions did not reveal any clear relation between lesion anatomy and segmentation performance across models lacking contrast-enhanced sequences (Figure 6), other than as expected with lesion size[36], suggesting broad invariance to spatially-defined anatomical features.

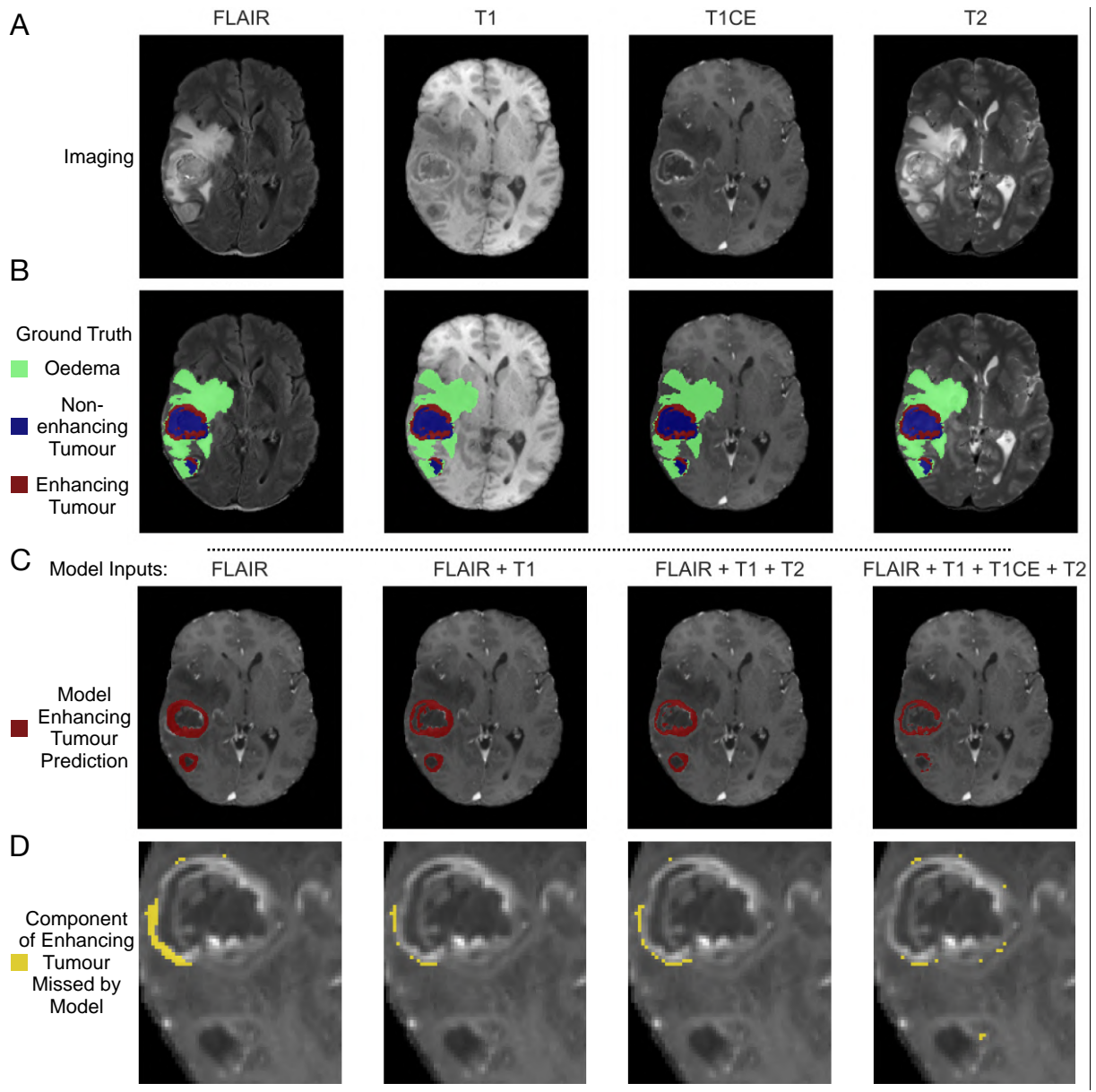

Figure 4: Segmenting enhancing tumour without contrast. A) Top panel illustrates axial slices of the lesion across the four sequences. B) Second panel illustrates the radiologist hand-labelled ground truth for the three tissue classes – of note red depicts enhancing tumour. C) Third panel illustrates predictions of enhancing tumour segmentation for four models with the following input data: i) FLAIR alone; ii) FLAIR and T1; iii) FLAIR, T1 and T2; and iv) FLAIR, T1, T2 and T1CE. Of note, only the final model is exposed to contrast-enhanced imaging, although the other three models still reasonably identify the location of the enhancing component. D) Fourth panel illustrates the component of enhancing tumour that is missed by the model.

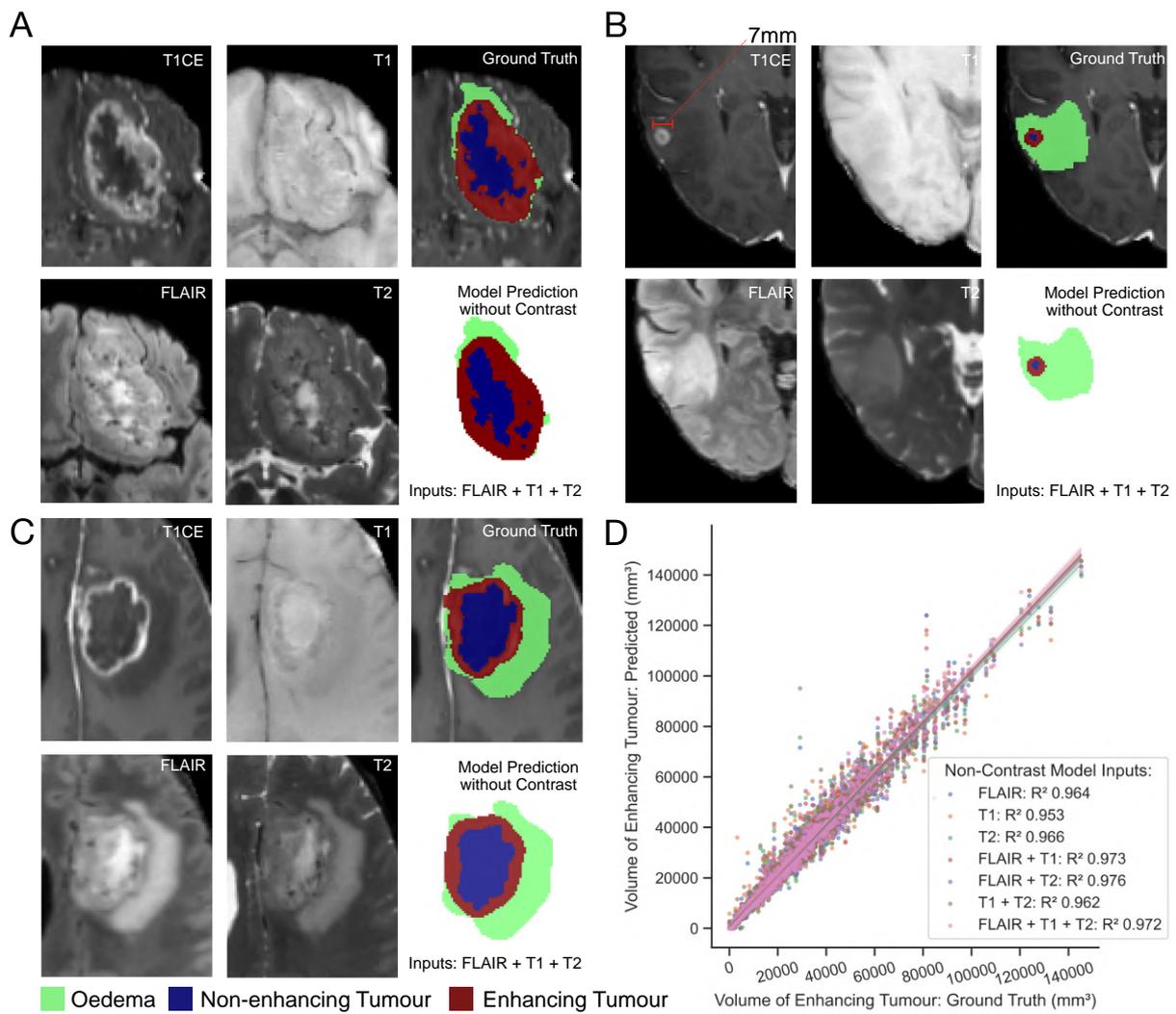

Figure 5: **Further examples of segmenting enhancing tumour without contrast.** A-C) Left two columns and rows of each panel illustrate the anatomical imaging for three randomly selected cases, whilst the third column of each panel illustrates the hand-labelled ground truth shown with the overlayed T1CE image, and finally the model prediction where contrast imaging was not provided. Of note, the case in panel B comprised a tumour with only a 7mm diameter enhancing component. D) The volume of enhancing tumour is highly significantly correlated to that of all model predictions, even when contrast-enhanced imaging is not provided (quantified by linear regression).

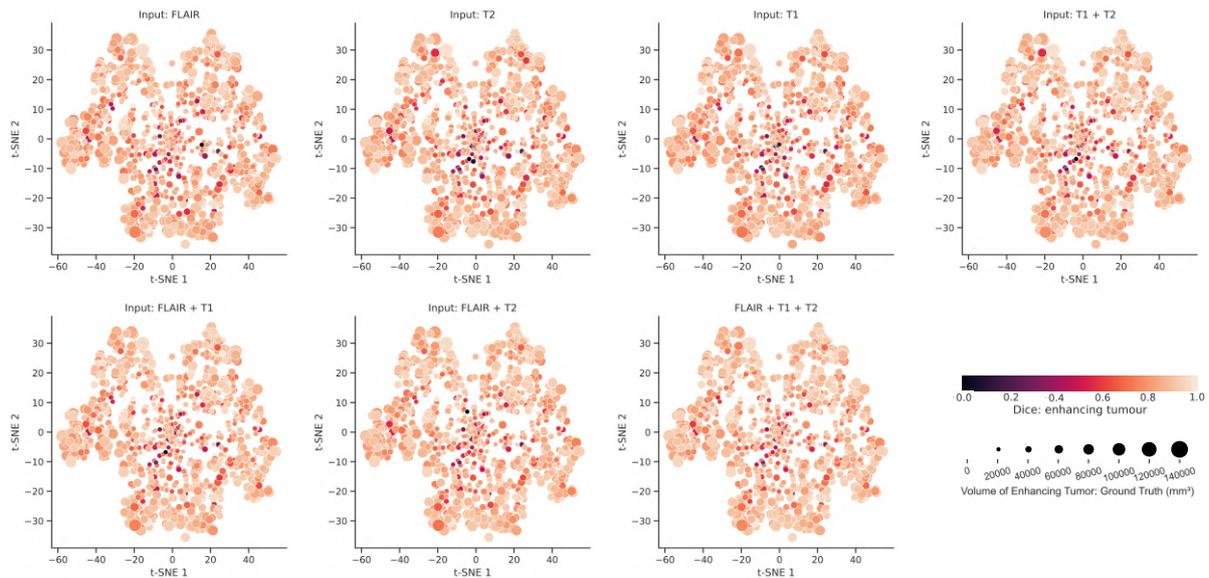

Figure 6: **Enhancing tumour segmentation is invariant to lesion morphology.** Two-dimensional t-SNE embeddings of the enhancing component of all lesions. Each panel illustrates the different set of non-contrast sequences used. Point size is proportional to enhancing tumour volume, whereas colour is proportional to Dice score. Point and colour keys are given at the bottom right of the plot. Note the expected lower scores for smaller segments, but no other obvious systematic variation across the latent space. The mathematical derivation of Dice coefficient and overview of tSNE is given in the methods.

### International clinical validation

*Whole tumour segmentation:* We evaluated the performance of all trained models on an out-of-sample cohort of 50 patients from our own centre in which lesions were hand-labelled, with scans acquired on both 1.5 and 3T scanners, and with a mixture of pre- and post-operative imaging. The cross-validation performances of all models from the BraTS data were well reproducible on our own data, with Dice coefficients for all models significantly correlated (r = 0.97, $p$ < 0.0001) (Figure 7). This was despite the multiple steps taken to deliberately make data more heterogenous and liable to error. As expected, models with single imaging modalities, such as T1 or T2 sequences alone, performed worst, with incremental gains in performance with alternative and supplementary modalities.

*Tissue class segmentation:* We manually reviewed the tissue segmentations of our own data predicted by the complete four-sequence model and determined the model's performances classifying tumour by subclasses of non-enhancing tumour, enhancing tumour and oedema were qualitatively more accurate than our semi-automated hand-segmentation. Akin to the method employed by the BraTS 2021 challenge[20], we therefore then utilized these model predictions using complete imaging sets as our new ground-truth with subsequent manual checking and refinement where required. We then compared the performance of all other models, i.e., those without four sequences, to this revised ground-truth. Model performances were again highly reproducible between the BraTS 2021 challenge data and that of our own external sample, with Dice coefficients significantly correlated (r = 0.95, $p$ < 0.0001) (Figure 7). As is usually the case in brain tumour segmentation models, segmentations for the non-enhancing tumour component fared worst – especially those with single imaging modalities, whilst prediction of enhancing tumour or oedema fared much better.

We applied our segmentation pipeline to a single patient from our own centre with variable quality (and availability) of imaging during their routine clinical care between 2010-2015. We also used this to quantitatively demonstrate lesion volumetry across this time, showing treatment response in early years, followed by stability, and later disease progression (Figure 8).

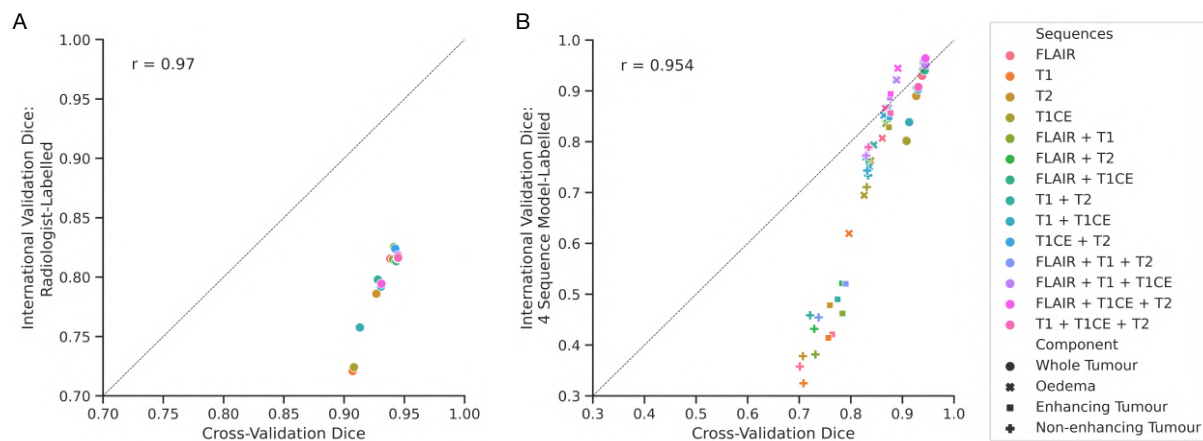

Figure 7: International validation. A) Scatterplot illustrates the strong relationship between radiologist-labelled lesions from a disparate international centre. Only relationship between whole tumour hand-segmentations and model predictions are shown here, as it transpired that the complete four sequence model more accurately delineated tissue classes than when hand-labelled. B) Scatterplot illustrates the strong relationship between model performances from the validation set, and when re-evaluated on our own data. For this plot, the complete four sequence model was utilized as the ground truth for the tissue subclasses of the international validation data. The mathematical derivation of the Dice coefficient is given in the methods.

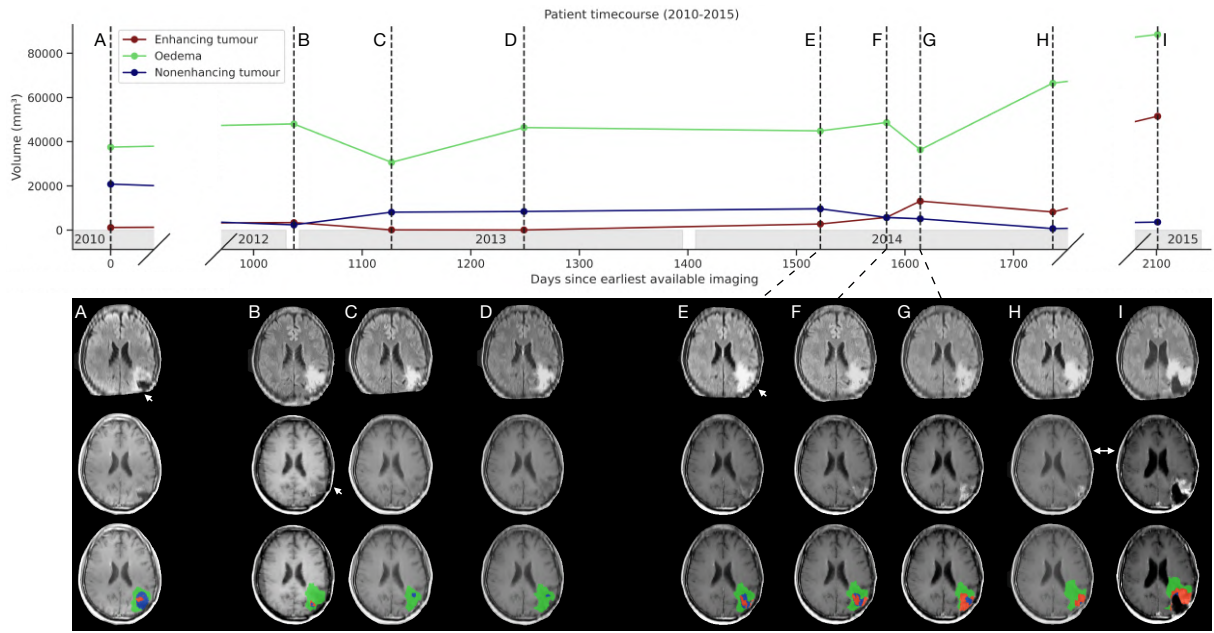

Figure 8: Single case example with longitudinal imaging between 2010 to 2015. Line-plot shows time on the x-axis (as days since earliest available imaging) and lesion compartmental volume determined from the segmentation model on the y-axis. Below are FLAIR (first row), T1CE (second row), and T1CE with predicted segmentation overlayed (third row) for each available scanning session. T1 and T2 images are not shown as were not available for all imaging sessions. Per the colour key, red depicts enhancing tumour, blue is non-enhancing tumour, and green is oedema. A) Mid-2010 imaging shows the FLAIR (originally coronal acquisition, but here reconstructed into axial with our super-resolution pipeline for visualisation), does not cover the posterior lesion entirely (white arrow). B) Early 2013 imaging shows the T1CE in some planes of acquisition did not fully cover the cortical surface (note the perfectly vertical line on either side of the brain cortical surface), and thus is super-resolved by using other sequences to resolve this (white arrow). E) Early 2014 FLAIR image demonstrates suboptimal image quality, and yet the segmentation model still delineates the tissue components subjectively well. H-I) T1CE images undertaken during late 2014 and 2015, respectively, show radical difference in image contrast, but that the segmentation model still performs subjectively well. Moreover, in panel H the model still recognises the surgical cavity not to be lesion, despite never being trained with post-operative imaging.

## Discussion

We have systematically surveyed the ability of state-of-the-art tumour segmentation models in delineating and quantifying brain tumour components in real-world clinical situations of incomplete and/or low-quality data. We reveal there is surprisingly little variation in the performance of segmenting a whole tumour with the number of modelled imaging modalities. Greater variation is observed when segmenting tumour components: a clear pattern of incremental improvement with the addition of further sequences emerges. These findings open the door both to the application of segmentation models to large-scale historical data, for the purpose of building treatment and outcome predictive models, and their deployment to real-world clinical care.

Strikingly, we find that segmentation models trained without contrast-enhancing imaging still characterize the anatomy of enhancing tumour components remarkably well. This includes quantification of the volumetric burden of enhancing tumour with high accuracy. Out-of-sample validation illustrates strong generalizability of these findings, including across super-resolved non-isotropic acquisitions, in varying MRI field strengths, and in tumour recurrences on complex postoperative imaging of limited quality. Our analyses show that current segmentation models generalize surprisingly well to real-world clinical imaging varying in quality and sequence completeness. We also use a case-based example (Figure 8) to demonstrate how this might factor into the clinical workflow, which in this case was achieved using a Docker container with Python, the software requirements as detailed in the methods, and the trained tumour segmentation model weights.

### Additive value of multiple sequences

Model fidelity unsurprisingly rose with the number of modelled sequences. What is, however, surprising is the ability of models based on limited data to delineate lesions very well. This is particularly striking in the segmentation of the whole tumour, where only marginal differences in Dice coefficient were seen across the range of sequence combinations. We can conclude that even single sequences may be sufficient for segmenting brain tumours with fidelity adequate for many downstream tasks.

The segmentation of tumour compartments—oedema, enhancing and non-enhancing tumour—however presents a more complex picture. Single sequence models of oedema and enhancing tumour perform best with FLAIR and T1CE sequences, respectively. But models of two or three sequences exhibit less intuitive behaviour. Adding FLAIR to T1CE achieves whole tumour performance very close to that of the complete, four sequence model, despite receiving only half the data. To that end, single T1CE MRI studies (such as in stealth imaging) may therefore benefit from the addition of a FLAIR sequence to enable more optimum visualization of the entire lesion to aid pre-operative planning. A two-sequence model of T1 & T1CE can delineate oedema well without the T2 or FLAIR typically used to identify it. Overall, these findings illustrate the ability of contemporary computer vision models to extract information from multiple sequences with greater efficiency than intuitive perception may suggest[37,38].

### Segmenting enhancing tumour without contrast-enhanced imaging

Strikingly, we found that models without the contrast enhancing sequence (T1CE) can still segment what has been hand-labelled by experienced neuroradiologists with full imaging datasets as the enhancing component of the tumour well, not least with performance largely invariant to the size, shape, and neuroanatomical location of the enhancing component. This introduces the possibility—across both research and clinical practice—to make approximate inferences about the anatomy of enhancing components without the use of contrast. Moreover, that a model can identify what has been termed the *'enhancing'* tumour[19,38], without any information about its enhancing properties, reveals the presence of non-intuitive imaging features that could render the enhancing component quantifiable without the use of contrast.

This challenges the current dogma of 'enhancing tumour', given a machine can identify it without the administration of intravenous agents ordinarily required to reveal it. Further investigation of this possibility is warranted, including the detectability of the presence of any degree of enhancement. These findings also illustrate a clinically important opportunity in oncological imaging when contrast enhanced imaging cannot be acquired, not least in situations of repeated follow-up where the over-use of contrast should ideally be limited, for example to minimize Gadolinium retention in paediatric patients. We note recent research on completing image sets synthetically may be fruitful in this domain[39-42], as well as a wider body of literature aiming to reduce the requirement for contrast[43].

## Limitations

In our systematic evaluation of the ability of deep learning models to identify brain tumours with varying degrees of sequence-incomplete data, we opted to use one single self-configuring architecture – nnU-Net[28,29]. The use of this software is well justified given its validated performance across many domains of medical imaging[30]. But segmentation models are a rapidly evolving field, and so it is possible that other architectures might perform differently, perhaps even superiorly, to that used here. It is however important to note that our aim was not to identify the definitive 'best' tumour segmentation model, but to quantify the impact of sequence completeness. Our aim is to determine how such models could perform in real-world clinical situations where 'perfect' data rarely exists, quantifying their appropriateness for translation to the clinical frontline. Furthermore, BraTS training data includes only preoperative imaging, yet it is plausible that much of the value in segmentation models may lie in longitudinal follow-up including that of postoperative resection appearances. Whilst we included a selection of postoperative imaging in our additional external validation, a more dedicated evaluation in the postoperative setting should form an area for future investigation.

## Conclusion

Automated segmentation models can characterize tumours in real-world clinical situations of incomplete imaging data remarkably well. Such models are also able to identify enhancing tumour *without* the use of contrast-enhanced imaging, potentially providing clinical guidance in circumstances where contrast administration is contraindicated or where its repeated use should be minimized. This opens the way to quantifying enhancing components without the administration of intravenous agents, not least invites a revision of the notion of tumour enhancement if the same information can be extracted without contrast. Its applicability includes not just prospective scenarios wherein a full scan may not be possible such as patients unable to receive intravenous contrast, but also applies to historical datasets where certain sequences might not have been acquired. Out-of-sample validation illustrates strong generalizability, across non-isotropic acquisitions and even on complex postoperative imaging where tumours have recurred. Translation of such models to the clinical frontline for response assessment – where complete data is a rarity – may be easier than hitherto believed.

**Funding information:**

JKR was supported by the Guarantors of Brain, the NHS Topol Digital Fellowship, the Medical Research Council (MR/X00046X/1), and the UCL CDT i4health. PN is supported by the Wellcome Trust (213038/Z/18/Z) and the UCLH NIHR Biomedical Research Centre. HH is supported by the UCLH NIHR Biomedical Research Centre.

**Competing interests:**

None to declare.

**Authorship:** Conceptualization: JR, HH, PN; Methodology: JR, SM, RG; Software: JR, SM; Validation: JR, HH; Formal analysis: JR, SM; Manuscript writing, reviewing, and editing: JR, SM, RG, HH, PN. All authors have been involved in the writing of the manuscript and have read and approved the final version.

# Brain tumour segmentation with incomplete imaging data

## Supplementary Material

| Sequences | Whole Tumour: Accuracy | Whole Tumour: Dice | Whole Tumour: False Discovery Rate | Whole Tumour: False Negative Rate | Whole Tumour: False Omission Rate | Whole Tumour: False Positive Rate | Whole Tumour: Negative Predictive Value | Whole Tumour: Precision | Whole Tumour: Recall |
|---|---|---|---|---|---|---|---|---|---|
| FLAIR | 0.998 (0.998-0.998) | 0.938 (0.936-0.941) | 0.055 (0.053-0.058) | 0.066 (0.063-0.069) | 0.001 (0.001-0.001) | 0.001 (0.001-0.001) | 0.999 (0.999-0.999) | 0.945 (0.942-0.947) | 0.934 (0.931-0.937) |
| T1 | 0.997 (0.997-0.997) | 0.907 (0.904-0.91) | 0.088 (0.085-0.091) | 0.096 (0.092-0.099) | 0.001 (0.001-0.002) | 0.001 (0.001-0.001) | 0.999 (0.998-0.999) | 0.912 (0.909-0.915) | 0.904 (0.901-0.908) |
| T2 | 0.998 (0.998-0.998) | 0.927 (0.925-0.929) | 0.069 (0.067-0.072) | 0.076 (0.073-0.078) | 0.001 (0.001-0.001) | 0.001 (0.001-0.001) | 0.999 (0.999-0.999) | 0.931 (0.928-0.933) | 0.924 (0.922-0.927) |
| T1CE | 0.997 (0.997-0.997) | 0.908 (0.905-0.911) | 0.088 (0.085-0.091) | 0.094 (0.091-0.097) | 0.001 (0.001-0.002) | 0.001 (0.001-0.001) | 0.999 (0.998-0.999) | 0.912 (0.909-0.915) | 0.906 (0.903-0.909) |
| FLAIR + T1 | 0.998 (0.998-0.998) | 0.941 (0.938-0.943) | 0.054 (0.052-0.057) | 0.063 (0.06-0.066) | 0.001 (0.001-0.001) | 0.001 (0.001-0.001) | 0.999 (0.999-0.999) | 0.946 (0.943-0.948) | 0.937 (0.934-0.94) |
| FLAIR + T2 | 0.998 (0.998-0.998) | 0.941 (0.939-0.944) | 0.053 (0.051-0.055) | 0.062 (0.059-0.065) | 0.001 (0.001-0.001) | 0.001 (0.001-0.001) | 0.999 (0.999-0.999) | 0.947 (0.945-0.949) | 0.938 (0.935-0.941) |
| FLAIR + T1CE | 0.998 (0.998-0.998) | 0.943 (0.941-0.946) | 0.051 (0.049-0.054) | 0.06 (0.057-0.063) | 0.001 (0.001-0.001) | 0.001 (0.001-0.001) | 0.999 (0.999-0.999) | 0.949 (0.946-0.951) | 0.94 (0.937-0.943) |
| T1 + T2 | 0.998 (0.998-0.998) | 0.928 (0.926-0.93) | 0.066 (0.064-0.068) | 0.076 (0.074-0.079) | 0.001 (0.001-0.001) | 0.001 (0.001-0.001) | 0.999 (0.999-0.999) | 0.934 (0.932-0.936) | 0.924 (0.921-0.926) |
| T1 + T1CE | 0.997 (0.997-0.997) | 0.913 (0.91-0.916) | 0.083 (0.08-0.085) | 0.089 (0.086-0.092) | 0.001 (0.001-0.001) | 0.001 (0.001-0.001) | 0.999 (0.999-0.999) | 0.917 (0.915-0.92) | 0.911 (0.908-0.914) |
| T1CE + T2 | 0.998 (0.998-0.998) | 0.931 (0.928-0.933) | 0.065 (0.063-0.068) | 0.072 (0.069-0.075) | 0.001 (0.001-0.001) | 0.001 (0.001-0.001) | 0.999 (0.999-0.999) | 0.935 (0.932-0.937) | 0.928 (0.925-0.931) |
| FLAIR + T1 + T2 | 0.998 (0.998-0.998) | 0.942 (0.94-0.945) | 0.054 (0.052-0.056) | 0.06 (0.057-0.062) | 0.001 (0.001-0.001) | 0.001 (0.001-0.001) | 0.999 (0.999-0.999) | 0.946 (0.944-0.948) | 0.94 (0.938-0.943) |
| FLAIR + T1 + T1CE | 0.998 (0.998-0.998) | 0.944 (0.942-0.946) | 0.051 (0.049-0.053) | 0.059 (0.056-0.061) | 0.001 (0.001-0.001) | 0.001 (0.001-0.001) | 0.999 (0.999-0.999) | 0.949 (0.947-0.951) | 0.941 (0.939-0.944) |
| FLAIR + T1CE + T2 | 0.998 (0.998-0.998) | 0.945 (0.943-0.947) | 0.049 (0.047-0.051) | 0.059 (0.057-0.062) | 0.001 (0.001-0.001) | 0.001 (0.001-0.001) | 0.999 (0.999-0.999) | 0.951 (0.949-0.953) | 0.941 (0.938-0.943) |
| T1 + T1CE + T2 | 0.998 (0.998-0.998) | 0.931 (0.929-0.933) | 0.066 (0.063-0.068) | 0.071 (0.068-0.074) | 0.001 (0.001-0.001) | 0.001 (0.001-0.001) | 0.999 (0.999-0.999) | 0.934 (0.932-0.937) | 0.929 (0.926-0.932) |
| FLAIR + T1 + T1CE + T2 | 0.998 (0.998-0.998) | 0.945 (0.943-0.947) | 0.05 (0.048-0.052) | 0.059 (0.056-0.062) | 0.001 (0.001-0.001) | 0.001 (0.001-0.001) | 0.999 (0.999-0.999) | 0.95 (0.948-0.952) | 0.941 (0.938-0.944) |

Supplementary Table 1: Detailed performance metrics of all sequence combinations across the models in segmenting whole tumour

| Sequences | Oedema: Accuracy | Oedema: Dice | Oedema: False Discovery Rate | Oedema: False Negative Rate | Oedema: False Omission Rate | Oedema: False Positive Rate | Oedema: Negative Predictive Value | Oedema: Precision | Oedema: Recall |
|---|---|---|---|---|---|---|---|---|---|
| FLAIR | 0.998 (0.998-0.998) | 0.861 (0.856-0.866) | 0.136 (0.13-0.141) | 0.137 (0.132-0.143) | 0.001 (0.001-0.001) | 0.001 (0.001-0.001) | 0.999 (0.999-0.999) | 0.864 (0.859-0.87) | 0.863 (0.857-0.868) |
| T1 | 0.997 (0.996-0.997) | 0.797 (0.79-0.803) | 0.199 (0.192-0.206) | 0.201 (0.195-0.206) | 0.002 (0.002-0.002) | 0.002 (0.002-0.002) | 0.998 (0.998-0.998) | 0.801 (0.794-0.808) | 0.799 (0.794-0.805) |
| T2 | 0.997 (0.997-0.997) | 0.839 (0.833-0.844) | 0.157 (0.151-0.163) | 0.161 (0.156-0.167) | 0.001 (0.001-0.001) | 0.001 (0.001-0.001) | 0.999 (0.999-0.999) | 0.843 (0.837-0.849) | 0.839 (0.833-0.844) |
| T1CE | 0.997 (0.997-0.997) | 0.826 (0.82-0.831) | 0.174 (0.168-0.18) | 0.169 (0.164-0.174) | 0.001 (0.001-0.002) | 0.001 (0.001-0.002) | 0.999 (0.998-0.999) | 0.826 (0.82-0.832) | 0.831 (0.826-0.836) |
| FLAIR + T1 | 0.998 (0.998-0.998) | 0.868 (0.863-0.873) | 0.126 (0.121-0.132) | 0.133 (0.128-0.138) | 0.001 (0.001-0.001) | 0.001 (0.001-0.001) | 0.999 (0.999-0.999) | 0.874 (0.868-0.879) | 0.867 (0.862-0.872) |
| FLAIR + T2 | 0.998 (0.998-0.998) | 0.871 (0.866-0.876) | 0.123 (0.118-0.128) | 0.131 (0.126-0.136) | 0.001 (0.001-0.001) | 0.001 (0.001-0.001) | 0.999 (0.999-0.999) | 0.877 (0.872-0.882) | 0.869 (0.864-0.874) |
| FLAIR + T1CE | 0.998 (0.998-0.998) | 0.888 (0.883-0.892) | 0.109 (0.104-0.114) | 0.112 (0.107-0.117) | 0.001 (0.001-0.001) | 0.001 (0.001-0.001) | 0.999 (0.999-0.999) | 0.891 (0.886-0.896) | 0.888 (0.883-0.893) |
| T1 + T2 | 0.997 (0.997-0.998) | 0.844 (0.839-0.85) | 0.145 (0.14-0.15) | 0.161 (0.156-0.167) | 0.001 (0.001-0.001) | 0.001 (0.001-0.001) | 0.999 (0.999-0.999) | 0.855 (0.85-0.86) | 0.839 (0.833-0.844) |
| T1 + T1CE | 0.997 (0.997-0.997) | 0.836 (0.831-0.841) | 0.166 (0.16-0.172) | 0.157 (0.153-0.162) | 0.001 (0.001-0.001) | 0.001 (0.001-0.001) | 0.999 (0.999-0.999) | 0.834 (0.828-0.84) | 0.843 (0.838-0.847) |
| T1CE + T2 | 0.998 (0.998-0.998) | 0.864 (0.859-0.869) | 0.133 (0.127-0.138) | 0.135 (0.13-0.14) | 0.001 (0.001-0.001) | 0.001 (0.001-0.001) | 0.999 (0.999-0.999) | 0.867 (0.862-0.873) | 0.865 (0.86-0.87) |
| FLAIR + T1 + T2 | 0.998 (0.998-0.998) | 0.873 (0.868-0.878) | 0.122 (0.117-0.127) | 0.129 (0.124-0.134) | 0.001 (0.001-0.001) | 0.001 (0.001-0.001) | 0.999 (0.999-0.999) | 0.878 (0.873-0.883) | 0.871 (0.866-0.876) |
| FLAIR + T1 + T1CE | 0.998 (0.998-0.998) | 0.889 (0.884-0.894) | 0.108 (0.103-0.113) | 0.11 (0.105-0.115) | 0.001 (0.001-0.001) | 0.001 (0.001-0.001) | 0.999 (0.999-0.999) | 0.892 (0.887-0.897) | 0.89 (0.885-0.895) |
| FLAIR + T1CE + T2 | 0.998 (0.998-0.998) | 0.891 (0.887-0.896) | 0.104 (0.099-0.109) | 0.109 (0.105-0.114) | 0.001 (0.001-0.001) | 0.001 (0.001-0.001) | 0.999 (0.999-0.999) | 0.896 (0.891-0.901) | 0.891 (0.886-0.895) |
| T1 + T1CE + T2 | 0.998 (0.998-0.998) | 0.867 (0.862-0.872) | 0.129 (0.124-0.134) | 0.133 (0.128-0.138) | 0.001 (0.001-0.001) | 0.001 (0.001-0.001) | 0.999 (0.999-0.999) | 0.871 (0.866-0.876) | 0.867 (0.862-0.872) |
| FLAIR + T1 + T1CE + T2 | 0.998 (0.998-0.998) | 0.891 (0.886-0.896) | 0.105 (0.1-0.11) | 0.11 (0.105-0.115) | 0.001 (0.001-0.001) | 0.001 (0.001-0.001) | 0.999 (0.999-0.999) | 0.895 (0.89-0.9) | 0.89 (0.885-0.895) |

Supplementary Table 2: Perilesional oedema detailed performance metrics of all sequence combinations

| Sequences | Enhancing Tumour: Accuracy | Enhancing Tumour: Dice | Enhancing Tumour: False Discovery Rate | Enhancing Tumour: False Negative Rate | Enhancing Tumour: False Omission Rate | Enhancing Tumour: False Positive Rate | Enhancing Tumour: Negative Predictive Value | Enhancing Tumour: Precision | Enhancing Tumour: Recall |
|---|---|---|---|---|---|---|---|---|---|
| FLAIR | 0.998 (0.998-0.999) | 0.764 (0.756-0.772) | 0.236 (0.228-0.244) | 0.216 (0.208-0.223) | 0.001 (0.001-0.001) | 0.001 (0.001-0.001) | 0.999 (0.999-0.999) | 0.764 (0.756-0.772) | 0.784 (0.777-0.792) |
| T1 | 0.998 (0.998-0.998) | 0.756 (0.748-0.765) | 0.25 (0.241-0.258) | 0.217 (0.21-0.224) | 0.001 (0.001-0.001) | 0.001 (0.001-0.001) | 0.999 (0.999-0.999) | 0.75 (0.742-0.759) | 0.783 (0.776-0.79) |
| T2 | 0.998 (0.998-0.999) | 0.759 (0.751-0.768) | 0.244 (0.235-0.253) | 0.215 (0.208-0.222) | 0.001 (0.001-0.001) | 0.001 (0.001-0.001) | 0.999 (0.999-0.999) | 0.756 (0.747-0.765) | 0.785 (0.778-0.792) |
| T1CE | 0.999 (0.999-0.999) | 0.874 (0.866-0.882) | 0.127 (0.118-0.135) | 0.106 (0.101-0.111) | 0.0 (0.0-0.0) | 0.0 (0.0-0.0) | 1.0 (1.0-1.0) | 0.873 (0.865-0.882) | 0.894 (0.889-0.899) |
| FLAIR + T1 | 0.999 (0.999-0.999) | 0.784 (0.776-0.792) | 0.222 (0.214-0.23) | 0.191 (0.185-0.198) | 0.001 (0.001-0.001) | 0.001 (0.001-0.001) | 0.999 (0.999-0.999) | 0.778 (0.77-0.786) | 0.809 (0.802-0.815) |
| FLAIR + T2 | 0.999 (0.999-0.999) | 0.783 (0.775-0.791) | 0.22 (0.211-0.228) | 0.194 (0.187-0.2) | 0.001 (0.001-0.001) | 0.001 (0.001-0.001) | 0.999 (0.999-0.999) | 0.78 (0.772-0.789) | 0.806 (0.8-0.813) |
| FLAIR + T1CE | 0.999 (0.999-0.999) | 0.875 (0.867-0.883) | 0.123 (0.115-0.131) | 0.106 (0.101-0.111) | 0.0 (0.0-0.0) | 0.0 (0.0-0.0) | 1.0 (1.0-1.0) | 0.877 (0.869-0.885) | 0.894 (0.889-0.899) |
| T1 + T2 | 0.999 (0.998-0.999) | 0.774 (0.766-0.782) | 0.23 (0.221-0.239) | 0.203 (0.196-0.21) | 0.001 (0.001-0.001) | 0.001 (0.001-0.001) | 0.999 (0.999-0.999) | 0.77 (0.761-0.779) | 0.797 (0.79-0.804) |
| T1 + T1CE | 0.999 (0.999-0.999) | 0.877 (0.87-0.885) | 0.119 (0.111-0.126) | 0.109 (0.104-0.113) | 0.0 (0.0-0.0) | 0.0 (0.0-0.0) | 1.0 (1.0-1.0) | 0.881 (0.874-0.889) | 0.891 (0.887-0.896) |
| T1CE + T2 | 0.999 (0.999-0.999) | 0.875 (0.867-0.883) | 0.123 (0.115-0.131) | 0.105 (0.1-0.11) | 0.0 (0.0-0.0) | 0.0 (0.0-0.0) | 1.0 (1.0-1.0) | 0.877 (0.869-0.885) | 0.895 (0.89-0.9) |
| FLAIR + T1 + T2 | 0.999 (0.999-0.999) | 0.79 (0.782-0.798) | 0.215 (0.207-0.224) | 0.185 (0.178-0.191) | 0.001 (0.001-0.001) | 0.001 (0.001-0.001) | 0.999 (0.999-0.999) | 0.785 (0.776-0.793) | 0.815 (0.809-0.822) |
| FLAIR + T1 + T1CE | 0.999 (0.999-0.999) | 0.875 (0.867-0.884) | 0.123 (0.115-0.132) | 0.105 (0.1-0.11) | 0.0 (0.0-0.0) | 0.0 (0.0-0.0) | 1.0 (1.0-1.0) | 0.877 (0.868-0.885) | 0.895 (0.89-0.9) |
| FLAIR + T1CE + T2 | 0.999 (0.999-0.999) | 0.877 (0.869-0.885) | 0.123 (0.115-0.131) | 0.104 (0.099-0.109) | 0.0 (0.0-0.0) | 0.0 (0.0-0.0) | 1.0 (1.0-1.0) | 0.877 (0.869-0.885) | 0.896 (0.891-0.901) |
| T1 + T1CE + T2 | 0.999 (0.999-0.999) | 0.877 (0.869-0.885) | 0.122 (0.114-0.131) | 0.104 (0.099-0.109) | 0.0 (0.0-0.0) | 0.0 (0.0-0.0) | 1.0 (1.0-1.0) | 0.878 (0.869-0.886) | 0.896 (0.891-0.901) |
| FLAIR + T1 + T1CE + T2 | 0.999 (0.999-0.999) | 0.876 (0.867-0.884) | 0.124 (0.116-0.133) | 0.103 (0.098-0.108) | 0.0 (0.0-0.0) | 0.0 (0.0-0.0) | 1.0 (1.0-1.0) | 0.876 (0.867-0.884) | 0.897 (0.892-0.902) |

Supplementary Table 3: Enhancing tumour detailed performance metrics of all sequence combinations

| Sequences | Non-enhancing Tumour: Accuracy | Non-enhancing Tumour: Dice | Non-enhancing Tumour: False Discovery Rate | Non-enhancing Tumour: False Negative Rate | Non-enhancing Tumour: False Omission Rate | Non-enhancing Tumour: False Positive Rate | Non-enhancing Tumour: Negative Predictive Value | Non-enhancing Tumour: Precision | Non-enhancing Tumour: Recall |
|---|---|---|---|---|---|---|---|---|---|
| FLAIR | 0.999 (0.999-0.999) | 0.701 (0.689-0.713) | 0.268 (0.256-0.28) | 0.297 (0.286-0.307) | 0.001 (0.0-0.001) | 0.0 (0.0-0.0) | 0.999 (0.999-1.0) | 0.732 (0.72-0.744) | 0.703 (0.693-0.714) |
| T1 | 0.999 (0.999-0.999) | 0.708 (0.697-0.72) | 0.262 (0.25-0.273) | 0.291 (0.28-0.302) | 0.001 (0.0-0.001) | 0.0 (0.0-0.0) | 0.999 (0.999-1.0) | 0.738 (0.727-0.75) | 0.709 (0.698-0.72) |
| T2 | 0.999 (0.999-0.999) | 0.707 (0.695-0.719) | 0.266 (0.255-0.278) | 0.29 (0.278-0.301) | 0.001 (0.0-0.001) | 0.0 (0.0-0.0) | 0.999 (0.999-1.0) | 0.734 (0.722-0.745) | 0.71 (0.699-0.722) |
| T1CE | 1.0 (0.999-1.0) | 0.831 (0.821-0.841) | 0.144 (0.134-0.154) | 0.166 (0.156-0.175) | 0.0 (0.0-0.0) | 0.0 (0.0-0.0) | 1.0 (1.0-1.0) | 0.856 (0.846-0.866) | 0.834 (0.825-0.844) |
| FLAIR + T1 | 0.999 (0.999-0.999) | 0.731 (0.72-0.743) | 0.241 (0.23-0.252) | 0.268 (0.257-0.279) | 0.0 (0.0-0.001) | 0.0 (0.0-0.0) | 1.0 (0.999-1.0) | 0.759 (0.748-0.77) | 0.732 (0.721-0.743) |
| FLAIR + T2 | 0.999 (0.999-0.999) | 0.729 (0.718-0.741) | 0.244 (0.233-0.256) | 0.262 (0.252-0.273) | 0.0 (0.0-0.0) | 0.0 (0.0-0.0) | 1.0 (1.0-1.0) | 0.756 (0.744-0.767) | 0.738 (0.727-0.748) |
| FLAIR + T1CE | 1.0 (0.999-1.0) | 0.831 (0.821-0.841) | 0.149 (0.139-0.159) | 0.16 (0.152-0.169) | 0.0 (0.0-0.0) | 0.0 (0.0-0.0) | 1.0 (1.0-1.0) | 0.851 (0.841-0.861) | 0.84 (0.831-0.848) |
| T1 + T2 | 0.999 (0.999-0.999) | 0.721 (0.709-0.733) | 0.252 (0.24-0.263) | 0.273 (0.262-0.284) | 0.0 (0.0-0.001) | 0.0 (0.0-0.0) | 1.0 (0.999-1.0) | 0.748 (0.737-0.76) | 0.727 (0.716-0.738) |
| T1 + T1CE | 1.0 (0.999-1.0) | 0.834 (0.824-0.844) | 0.147 (0.137-0.157) | 0.16 (0.152-0.169) | 0.0 (0.0-0.0) | 0.0 (0.0-0.0) | 1.0 (1.0-1.0) | 0.853 (0.843-0.863) | 0.84 (0.831-0.848) |
| T1CE + T2 | 1.0 (0.999-1.0) | 0.832 (0.821-0.842) | 0.144 (0.134-0.154) | 0.164 (0.155-0.173) | 0.0 (0.0-0.0) | 0.0 (0.0-0.0) | 1.0 (1.0-1.0) | 0.856 (0.846-0.866) | 0.836 (0.827-0.845) |
| FLAIR + T1 + T2 | 0.999 (0.999-0.999) | 0.738 (0.726-0.749) | 0.233 (0.222-0.244) | 0.259 (0.249-0.269) | 0.0 (0.0-0.0) | 0.0 (0.0-0.0) | 1.0 (1.0-1.0) | 0.767 (0.756-0.778) | 0.741 (0.731-0.751) |
| FLAIR + T1 + T1CE | 1.0 (0.999-1.0) | 0.83 (0.819-0.84) | 0.144 (0.134-0.153) | 0.167 (0.158-0.176) | 0.0 (0.0-0.0) | 0.0 (0.0-0.0) | 1.0 (1.0-1.0) | 0.856 (0.847-0.866) | 0.833 (0.824-0.842) |
| FLAIR + T1CE + T2 | 1.0 (0.999-1.0) | 0.834 (0.824-0.844) | 0.144 (0.134-0.154) | 0.162 (0.153-0.17) | 0.0 (0.0-0.0) | 0.0 (0.0-0.0) | 1.0 (1.0-1.0) | 0.856 (0.846-0.866) | 0.838 (0.83-0.847) |
| T1 + T1CE + T2 | 1.0 (0.999-1.0) | 0.834 (0.824-0.844) | 0.147 (0.137-0.156) | 0.158 (0.149-0.166) | 0.0 (0.0-0.0) | 0.0 (0.0-0.0) | 1.0 (1.0-1.0) | 0.853 (0.844-0.863) | 0.842 (0.834-0.851) |
| FLAIR + T1 + T1CE + T2 | 1.0 (0.999-1.0) | 0.826 (0.815-0.837) | 0.148 (0.138-0.158) | 0.166 (0.157-0.175) | 0.0 (0.0-0.0) | 0.0 (0.0-0.0) | 1.0 (1.0-1.0) | 0.852 (0.842-0.862) | 0.834 (0.825-0.843) |

Supplementary Table 4: Non-enhancing tumour detailed performance metrics of all sequence combinations